\documentclass[runningheads]{llncs}

\usepackage[mobile]{eccv}

\usepackage{eccvabbrv}

\usepackage{graphicx}
\usepackage{booktabs}
\usepackage[percent]{overpic}

\usepackage[accsupp]{axessibility}  

\usepackage{enumitem}
\setlist[itemize]{noitemsep, topsep=2pt}

\usepackage{hyperref}

\newcommand{\benchmarkname}{neuralCAD-Edit}

\newcommand{\easy}{\textcolor[rgb]{0.35,0.64,0.36}}
\newcommand{\medium}{\textcolor[rgb]{0.89,0.64,0.25}}
\newcommand{\hard}{\textcolor[rgb]{0.85,0.39,0.39}}
\newcommand{\baseline}{\textcolor[rgb]{0.44,0.44,0.44}}
\newcommand{\requestor}{\textcolor[rgb]{0.58,0.58,0.58}}
\newcommand{\claude}{\textcolor[rgb]{0.81,0.57,0.49}}
\newcommand{\gemini}{\textcolor[rgb]{0.46,0.61,0.89}}
\newcommand{\gpt}{\textcolor[rgb]{0.43,0.70,0.63}}

\begin{document}

\title{\benchmarkname: An Expert Benchmark for Multimodal-Instructed 3D CAD Model Editing}

\titlerunning{\benchmarkname}

\author{
Toby Perrett$^\dagger$, 
Matthew Bouchard, 
William McCarthy$^\dagger$
}

\authorrunning{T. Perrett et al.}

\institute{Autodesk Research \hspace{2pt}
\email{\{first\}.\{last\}@autodesk.com} \hspace{2pt} $^\dagger$joint project lead
}
\maketitle

\vspace{-15pt}
\begin{abstract}
  
We introduce \benchmarkname, the first benchmark for editing 3D CAD models collected from expert CAD engineers. Instead of text conditioning as in prior works, we collect realistic CAD editing requests by capturing videos of professional designers, interacting directly with CAD models in CAD software, while talking, pointing and drawing. We recruited ten consenting designers to contribute to this contained study. We benchmark leading foundation models against human CAD experts carrying out edits, and find a large performance gap in both automatic metrics and human evaluations. Even the best foundation model (GPT 5.2) scores 53\% lower (absolute) than CAD experts in human acceptance trials, demonstrating the challenge of \benchmarkname. We hope \benchmarkname\ will provide a solid foundation against which 3D CAD editing approaches and foundation models can be developed. Code/data: \url{https://autodeskailab.github.io/neuralCAD-Edit}

\end{abstract}

\begin{figure}[!b]
\vspace{-20pt}
    \centering
    \includegraphics[width=1\linewidth]{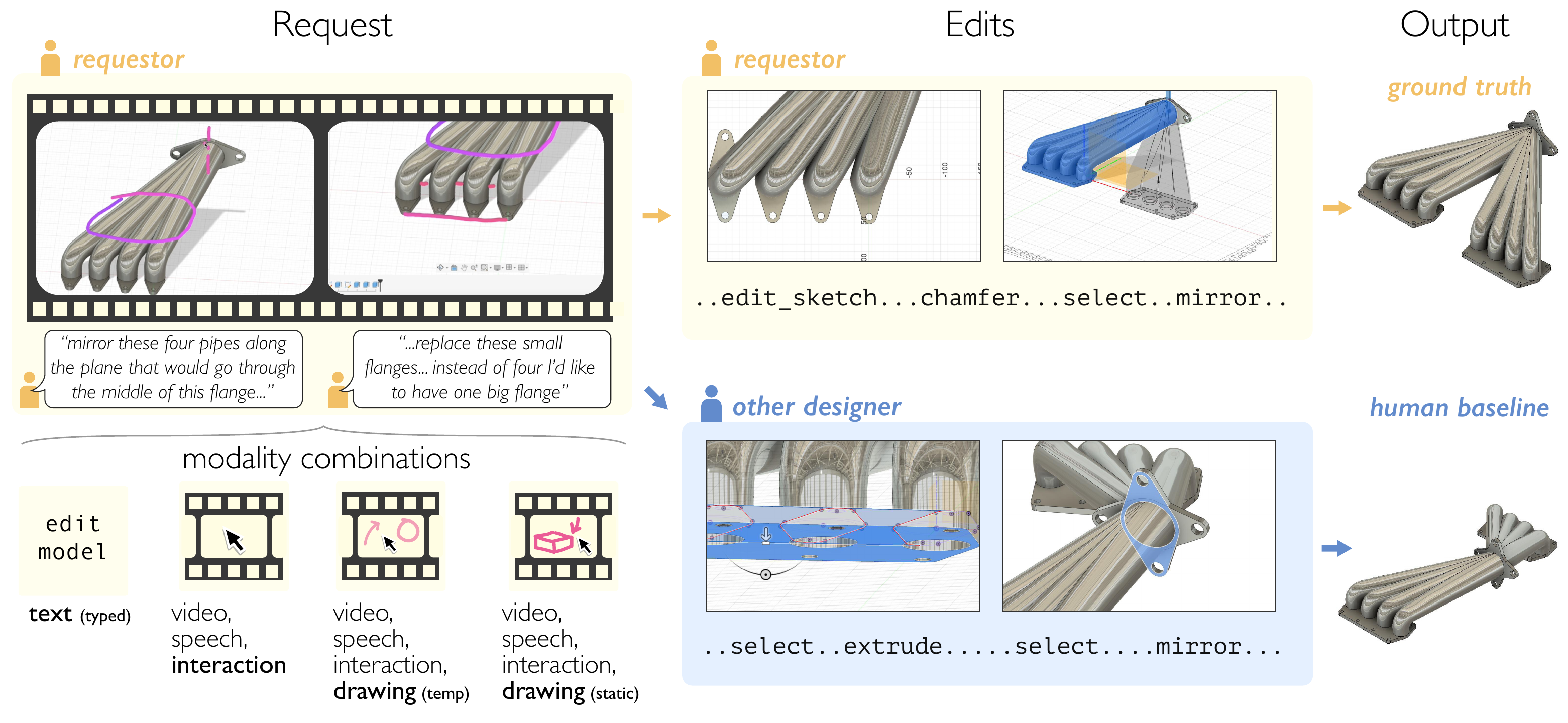}
    \vspace{-10pt}
    \caption{
    A human CAD expert is given an input 3D CAD model, and gives a multimodal edit \textbf{Request} whilst interacting with the model. \textbf{Edits} are then carried out by the requestor, and an additional CAD expert, to produce \textbf{Output} CAD models. This example highlights that even precise requests can contain some ambiguity. \benchmarkname\ captures four modality combinations for requests, and benchmarks the ability of AI approaches to understand multimodal requests and perform 3D edits against the standard set by human CAD experts. 
    }
    \label{fig:1}
\end{figure}

3D Computer Aided Design (CAD) generation is progressing rapidly \cite{wu2024cadvlm,xu2024brepgen,khan2024text2cad,xu2025autobrepautoregressivebrepgeneration,wang2025text,Chen_2025_CVPR,li2025caddreamer}.
The next frontier is editing. Editing is a crucial part of the design process, where CAD models are refined to meet tolerances and to repair errors, modified based on feedback from collaborators, and updated to reflect changes in design specifications.
Introducing a 3D CAD editing benchmark built around expert CAD user requests is the purpose of this paper.

3D CAD editing is an ideal task on which to evaluate frontier models' 3D spatio-temporal capabilities. It tests their ability to understand complex domain-specific audio-visual instructions and human design intent. In addition to this passive understanding of instructions, it also tests their ability to actively perform 3D geometry manipulation. These capabilities will be necessary to progress past the limitations of a ``text-box-only'' interface to something more intuitive, immersive, creative and productive.

While editing tasks in standard computer vision fields have focused on text \cite{brooks2023instructpix2pix,parelli20253d} or image \cite{couairon2022diffedit,li2025meshpad} conditioning, engineers who use 3D CAD software are capable of delivering much richer input to the programs they use.
Designers communicate with their colleagues using visual media (e.g. by sketching) \cite{williams2002designers, mccarthy2025mrcad}, but also gesture in visual space (e.g. gesturing to or on top of drawings) \cite{chandrasegaran2018sketching} while manipulating the CAD model over time.
This allows them to communicate spatial concepts more naturally, and to deliver instructions with the precision necessary for their professional environment.
For AI approaches to be naturally integrated into and improve designers' workflows, understanding complex temporally-interleaved multimodal editing instructions is vital \cite{williams2002designers}. However, prior CAD approaches have stuck to text conditioning only \cite{liu2025b,yuan2025cad}. In this paper, we allow designers to talk, point and draw their editing instructions in real-time, all with full simultaneous interactivity with the 3D model in a CAD environment. This is how they would naturally instruct a co-worker to do a task.

\Cref{fig:1} displays an example input CAD model, request and ground truth edit introduced in \benchmarkname, highlighting the complexity of the request. It also displays an edit performed by a different CAD expert, showing that even precise multimodal requests can contain some ambiguity that must be handled.

Our main contributions are:
\vspace{-2pt}
\begin{itemize}
\item We introduce \benchmarkname, the first dataset and benchmark for 3D CAD model editing collected from expert CAD engineers using professional CAD software. Ten designers consented and contributed to this controlled study.
\item \benchmarkname\ is the first 3D CAD editing benchmark to contain multimodal requests, featuring simultaneous and interleaved model manipulation, speech, drawing and video.
\item We perform a detailed analysis of designers' interactions and behavior.
We find that their multimodal instructions convey more information in less time, and elicit longer, more complex edits.
\item We benchmark leading foundation models against expert human CAD users, finding a large performance gap with even the best model (GPT 5.2), and explore 3D and feature-based metrics, human expert ratings and VLM evaluations.
\end{itemize}

\section{Related work}

\noindent\textbf{3D CAD generation and editing methods.}
CAD presents a number of additional challenges compared to other forms of 3D data. In contrast to other modalities including voxels \cite{zhou20213d}, point clouds \cite{luo2021diffusion}, meshes \cite{chen2025sam}, radiance fields \cite{mildenhall2020nerf} and splats \cite{kerbl20233d}, the industry-standard CAD modality, the Boundary Representation (B-Rep) \cite{weiler1986topological}, encourages easy editability of its vertices, edges and faces (the minimal number required, in contrast to a triangular mesh), and downstream manufacturing necessitates high quality outputs with a much lower tolerance for errors. We use B-Reps for all of our data.
General 3D generators and editors do not produce B-Reps \cite{li2025triposg,xiang2025structured,parelli20253d,li2025meshpad}.

Models for generating 3D B-Reps have made fast progress in recent years, following generation in other fields, incorporating diffusion \cite{wang2024vq,xu2024brepgen,lee2025brepdiff}, auto-regressive generation \cite{xu2025autobrepautoregressivebrepgeneration,li2025brepgpt}, and text- \cite{khan2024text2cad,wang2025text} and image- conditioning \cite{wu2024cadvlm,manvideocad,Chen_2025_CVPR,li2025caddreamer}. However, currently these only operate on a single body at a time, and they cannot modify models after generation to fit the designer's precise needs.

Some works are starting to look at specialised models for text-conditioned single component edits \cite{yuan2025cad,liu2025b}, and the addition of tool use and the emergent capabilities of general auto-regressive models is a promising direction. For example, CAD-Assistant \cite{mallis2025cad} proposes a tool-use LLM and evaluates on VQA and sketch auto-constraining and parameterisation. Researchers are also starting to provide harnesses for foundation models to write and execute CAD scripts \cite{xie2025text,badagabettu2024query2cad,Li_2025_CVPR,zuo2025cadhllm,yu2025gencad}. There is potential for all these approaches to extend to multimodal-conditioned edits on full assemblies in the future, and \benchmarkname\ aims to provide a benchmark for when they are ready.

\noindent\textbf{Non-CAD 3D editing benchmarks} are still rare as the community is focused on generation. 3D-LATTE \cite{parelli20253d} evaluates on 100 text-only requests for splat edits, and does not provide any human edits. BlenderGym \cite{gu2025blendergym} contains 50 geometry editing tasks. These are conditioned by a rendered image of the "goal", implying that a completed edit needs to exist, which does not reflect real world usage. Meshpad \cite{li2025meshpad} evaluates only on synthetic sketch conditioning.

\noindent\textbf{CAD 3D editing benchmarks.}
Most benchmarks for 3D CAD have been for generation. For example, VideoCAD \cite{manvideocad} starts with a series of operations (sketch, extrude etc.) which generates a B-Rep, and replays this through CAD software UI. However, it is only designed for and evaluated on image-conditioned generation and VQA.
CAD-Editor \cite{yuan2025cad} includes a synthetic benchmark for text-only single component editing. It generates edits with a ``next-command'' auto-complete, and relies on VLM captioning of edits to produce requests.
In contrast to these works, \benchmarkname\ uses human experts to request edits and perform them, on single components and assemblies, and measures success on the final edited model.

In contrast to all the above benchmarks (both non-CAD and CAD), \benchmarkname\ is the first to collect video and audio of experts giving multimodal instructions naturally. We record them interacting with the CAD model, whilst talking, pointing and drawing. 
For an AI to do well, it must understand this complex multimodal input, and handle the additional ambiguity in natural communication to understand the designer's intent. It must then possess sufficient 3D understanding to manipulate CAD geometry to a designer's exacting standards.

\noindent\textbf{Evaluation-only datasets}
The rise of learning-based methods for computer vision, and the desire to evaluate every method under exactly the same data regime,
led to datasets being released with train/val/test splits \cite{lecun2002gradient,deng2009imagenet,carreira2017quo}. However, the trend is now back towards smaller high-quality evaluation-only datasets, decoupling data used for training and benchmarking.
This is because (1) models are now pre-trained on internet-scale data \cite{radford2021learning,schuhmann2022laion}, (2) the community is coalescing around general models over specialist \cite{vaswani2017attention}, and (3) many innovations which make a practical difference are now in large-scale training data collection, generation \cite{Black_2023_CVPR} and annotation \cite{kirillov2023segment}, and training recipes \cite{deitke2024molmo,guo2025deepseek}, rather than method innovations. This is observed especially in benchmarks for fields which require 3D and multi-modal understanding, such as visual perception \cite{patraucean2023perception}, video understanding \cite{perrett2025hd}, embodied AI and robotics \cite{majumdar2024openeqa}, computer use \cite{agrawal2025uinavbench}, and agentic workflows which require experts to annotate or generate evaluation tasks \cite{phan2025humanity,patwardhan2025gdpval}. Most notable are examples which collect their own data 
and pair this with high quality human generated/reviewed annotations \cite{perrett2025hd,agrawal2025uinavbench,patwardhan2025gdpval}. We follow this principle for \benchmarkname.

\section{The \benchmarkname\ benchmark}

\benchmarkname\ is designed to benchmark the ability of AI\footnote{We use the term AI throughout to refer to machine learning approaches, so as not to cause confusion with ``model'', which refers to 3D CAD models.} to edit 3D CAD models in response to naturalistic requests. \emph{Requests} are collected from ten expert consenting CAD users recruited for this study, interacting with existing CAD models in production CAD software using speech, text, drawing and model manipulation and interaction. Importantly, the expert designers make their requests as if they are asking a colleague to carry out the work as part of their daily professional activity. \emph{Edits} are then performed in response to requests, by expert CAD users, producing a final edited CAD model.

\begin{figure}
    \centering
    \includegraphics[width=1\linewidth]{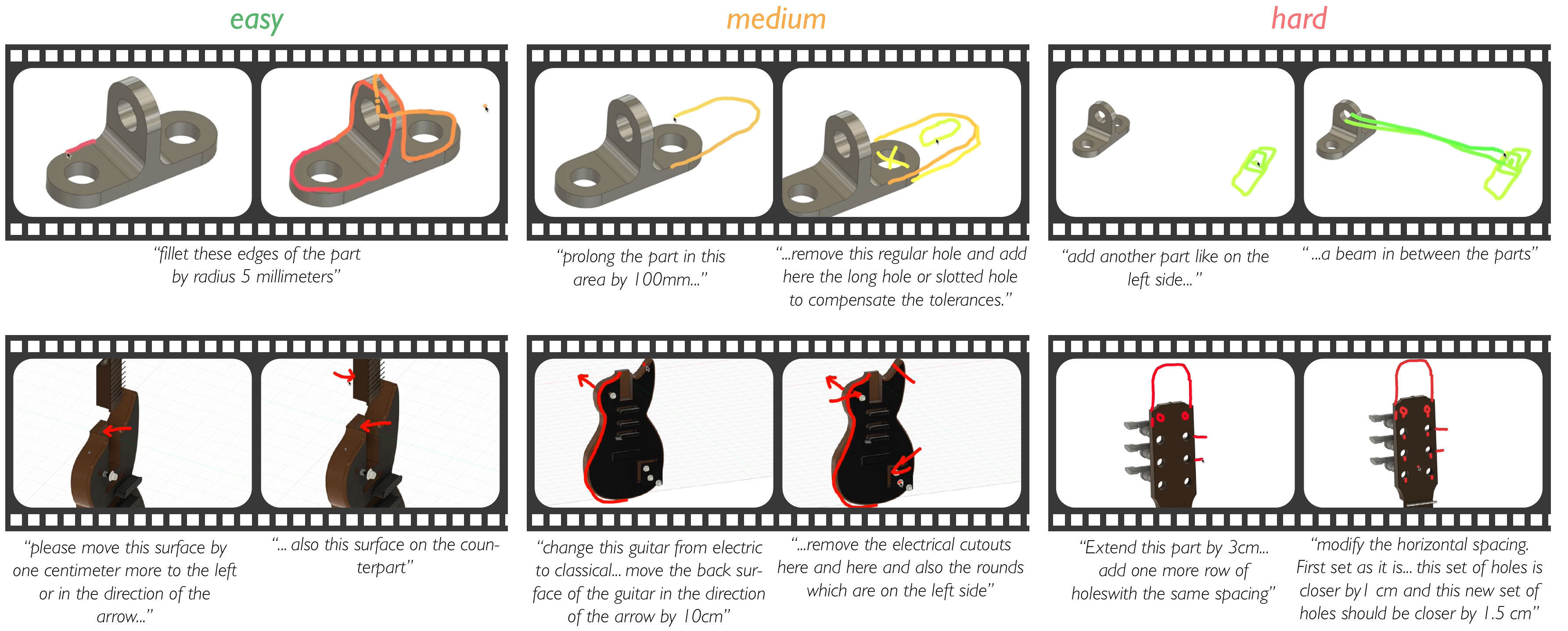}  \vspace{-15pt}
    \caption{For a given CAD model, experts requested edits that they thought would take professional designers 2 (easy), 5 (medium), and 10 (hard) minutes to complete. Video frames with transcribed speech snippets from the request are shown. Both simple (top) and complex (bottom) models can be the source of simple and complex edits.}
    \label{fig:request_difficulty}
\end{figure}

\subsection{Collection protocol}

In order to include a range of requests with short to long time horizons, for a given CAD model, the requestor is instructed to provide three independent requests - easy, medium and hard, expected to take 2, 5 and 10 minutes respectively.
Each row in \Cref{fig:request_difficulty} corresponds to a single initial CAD model, and shows associated easy, medium and hard requests with summarized descriptions. It highlights the variety of requests that can be made from a single model, and that even a simple model (top) can elicit a range of request difficulties.

For each request, we collect two edits. The first is from the original requestor. The second is from a different expert, who has seen the request video and input model (without seeing the original requestor's edited CAD model). 
This additional edit provides a human baseline.

All requests and edits are captured whilst participants are using Autodesk Fusion. A custom plugin collects the following data for requests: screen recording video (containing mouse, software UI and drawing), speech, selections. The following data is collected for edits: viewport renders, selections (e.g. which edge is clicked on), events (e.g. chamfer command), start and stop timestamps, final model and model state throughout the edit. Although our benchmark only relies on the input CAD model, request video/text and the final CAD model, we use this additional data to profile the dataset, and we will be releasing it all to encourage further research into 3D CAD editing.

\subsection{Modalities}

\begin{figure}[t!]
    \centering
    \includegraphics[width=1\linewidth]{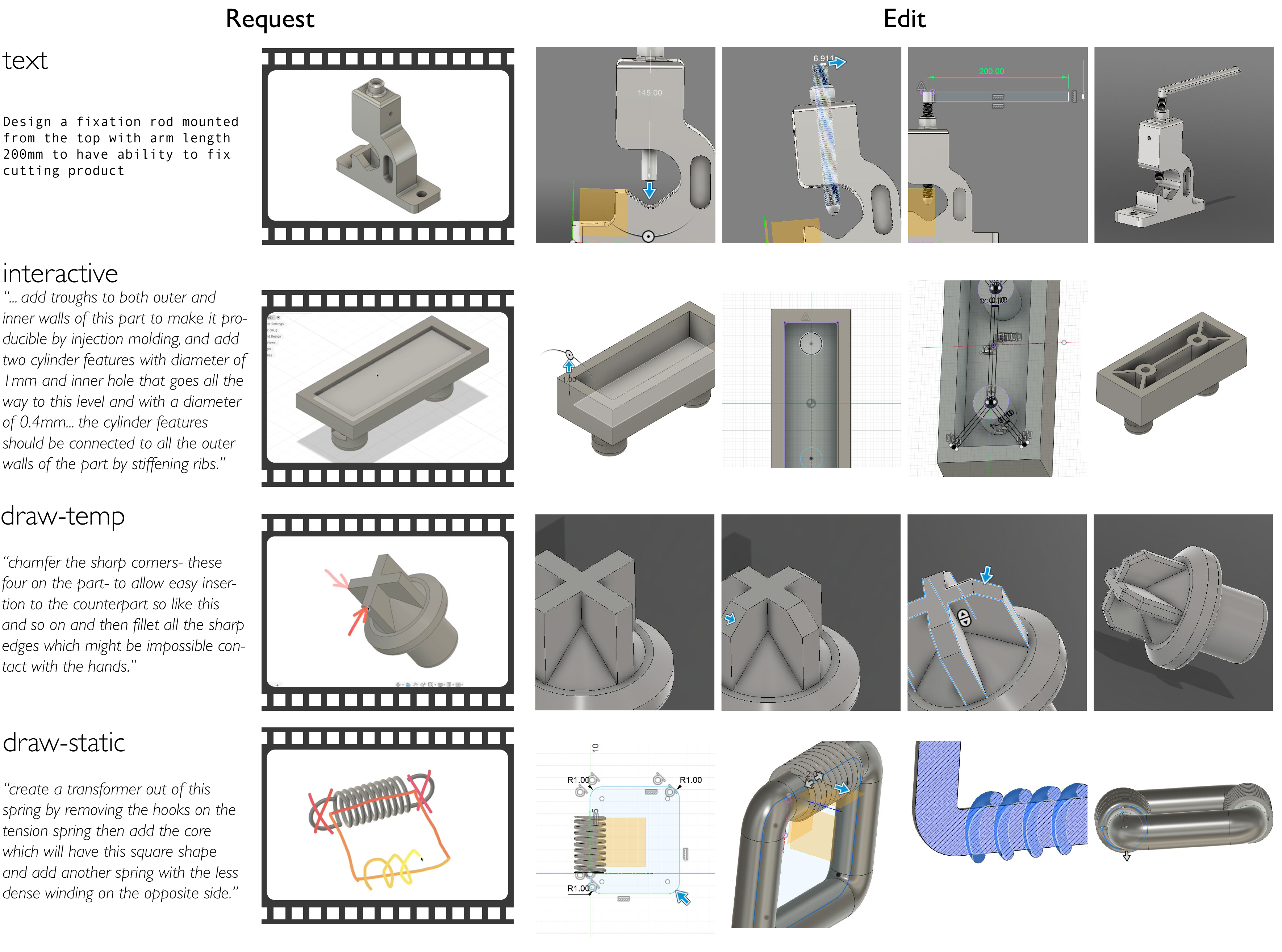}  \vspace{-20pt}
    \caption{Example requests from each modality and selected viewport renders from the associated human edits. The text request shows the typed text and the input model. The three video modalities show snippets of transcribed speech and one video frame.}
    \label{fig:edit_gallery}
\end{figure}

We collect requests expressed through the following modalities, with one example of each request and the associated edit shown in \Cref{fig:edit_gallery}

\begin{itemize}
\item \textbf{Text}. This is the simplest form of request, collected by typing into a text box with no model manipulation or speech. Text allows us to diagnose whether current limitations are due to multimodal instruction understanding (\ie video, speech and drawing), or 3D CAD model manipulation.
\item \textbf{Interactive}. Video of the requestor interacting with the CAD model, gesturing/pointing with the mouse, and speaking instructions concurrently.
\item \textbf{Draw-temp}. As interactive, but the requestor also uses a drawing tool to help illustrate the request. Drawings disappear after a few seconds.
\item \textbf{Draw-static}. As for draw-temp, but drawings remain visible for the whole request or until manually deleted.
\end{itemize}
Note that the video captures the drawing interactions and viewpoint changes directly as the requestor performs them. This maintains temporal alignment between interaction, drawing, speech and mouse movement.

\vspace{-20pt}
\subsection{Input CAD models}
\vspace{-2pt}

In order to capture editing on the different CAD model types designers encounter, we consider two input axes: (1) Parameterisation: whether a model is  \textit{dumb} or \textit{parametric} and (2) Construction: whether a model is a \textit{single-body} or multiple bodies in an \textit{assembly}. 
\textit{Parametric} models are those with a design history or set of configurations. In contrast, \textit{dumb} models are at their current construction state only, and would typically be encountered either as the output of a generative model, or an import of a file without any parametric history. \textit{Single-body} models contain a single solid, and are the current extent of what can be reliably generated by 3D CAD generators.
\textit{Assemblies} contain multiple single-bodies, and will be targeted by generative models in the near future.
\benchmarkname\ is balanced, with each Parameterisation/Construction combination contributing equally.
CAD models were sourced from the Fusion360 Gallery Dataset \cite{willis2020fusion, willis2022joinable}. All models were subject to human curation to filter designs containing personally identifying information or intellectual property.

\subsection{Validation, annotation and post-processing}
Each request and edit is manually checked by an expert CAD user. Those deemed not up to standard (12\%) are rejected. Request videos are extracted at 720p and 30FPS to keep file-sizes sensible. Speech is automatically transcribed by Whisper \cite{radford2023robust}. In addition to these automatic transcripts, we also provide oracle transcriptions - automatic transcriptions are manually checked, with errors corrected by the original requestor. This process repeats until the transcript is accepted by a different expert. 49\% of requests required at least one transcription correction, although most are minor, with an average character-wise edit distance of 3.2 against an average transcription length of 82.7 characters. For both automatic and oracle transcriptions, WhisperX \cite{bain2022whisperx} alignment is run on the transcripts to obtain timestamps for every word, allowing downstream models to associate \eg ``fillet this hole'' with the hole the mouse is hovering over at that time.

\vspace{-1pt}
\section{\benchmarkname\ data properties}
\vspace{-1pt}

\benchmarkname\ covers each of the 4 modalities, 4 parameterisations and 3 difficulties with 4 requests. Each request is carried out by the requestor and a different expert CAD user. In total, this is 192 requests (totaling 1.9 hours) and 384 edits (totaling 28.4 hours), 
produced by 10 expert CAD users with between 8 and 13 years of mechanical engineering design experience. 
In total 30K events are logged (with 24K associated viewport renders), with an average of 78 events per edit. 
Note that there is always a trade-off between dataset size and the feasibility/expense of running evaluations. As \benchmarkname\ is designed to be used for evaluation, this size strikes a good balance between coverage of the different modalities and the reliability of results, versus the ability to run large models on all requests, especially for academic labs\footnote{Current frontier models cost around \$200 to run on the whole benchmark.}. As a comparison, the evaluation set recently introduced for text-only non-CAD 3D editing in \cite{parelli20253d} contains 100 requests (and no human edits). Other task-based UI/interaction/agentic benchmarks adopt a similar approach. E.g. UINavBench \cite{agrawal2025uinavbench} contains 145 tasks and GDPval \cite{patwardhan2025gdpval} contains 5 tasks for each of its 44 occupations in its public subset.

\begin{figure}[t]
    \centering
    \begin{subfigure}{0.24\linewidth}
        \centering
        \includegraphics[height=82pt]{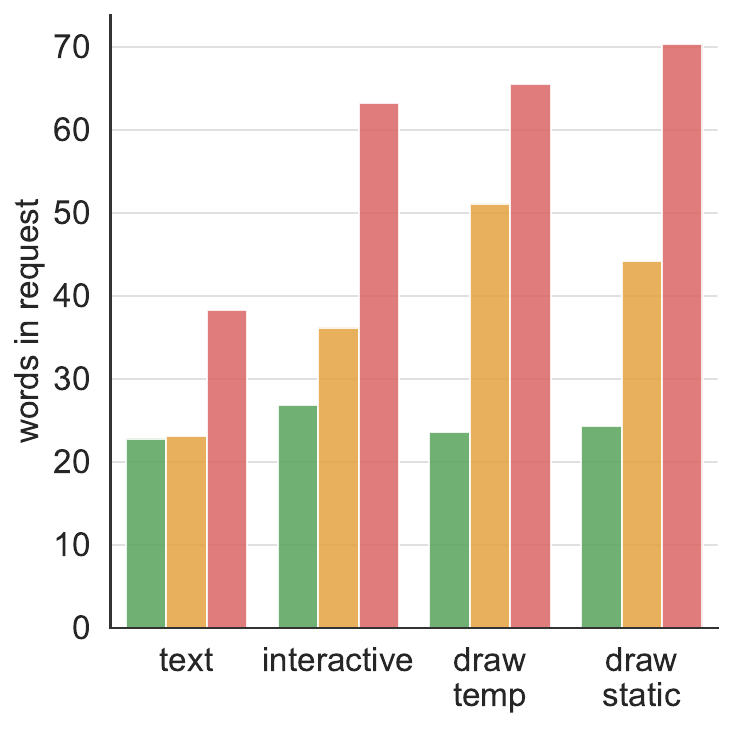}
        \caption{Number of words in request.}
    \label{fig:stats_requests_words}
    \end{subfigure}
\hfill
    \begin{subfigure}{0.24\linewidth}
        \centering
        \includegraphics[height=82pt]{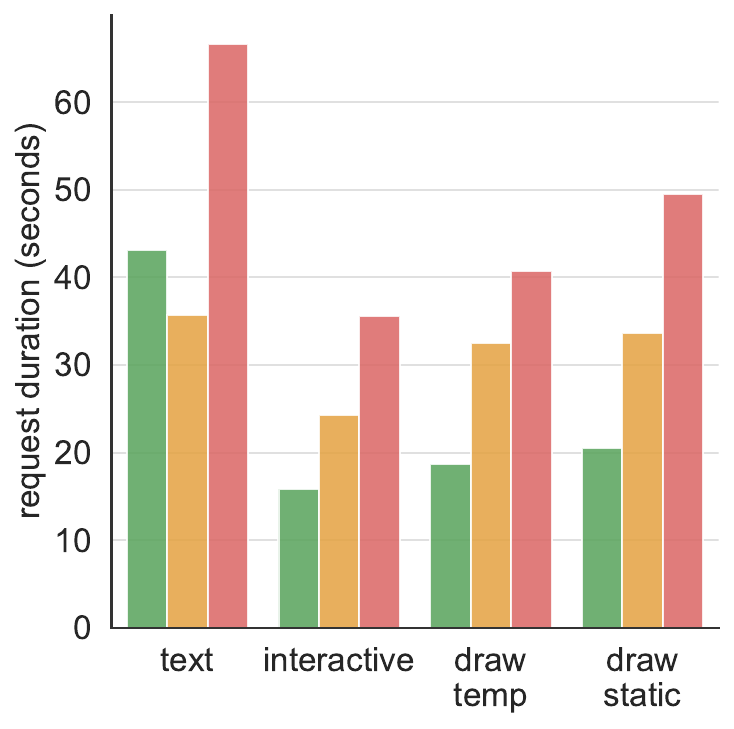}
        \caption{Time taken to produce request.}
        \label{fig:stats_requests_duration}
    \end{subfigure}
\hfill
    \begin{subfigure}{0.24\linewidth}
        \centering
        \includegraphics[height=82pt]{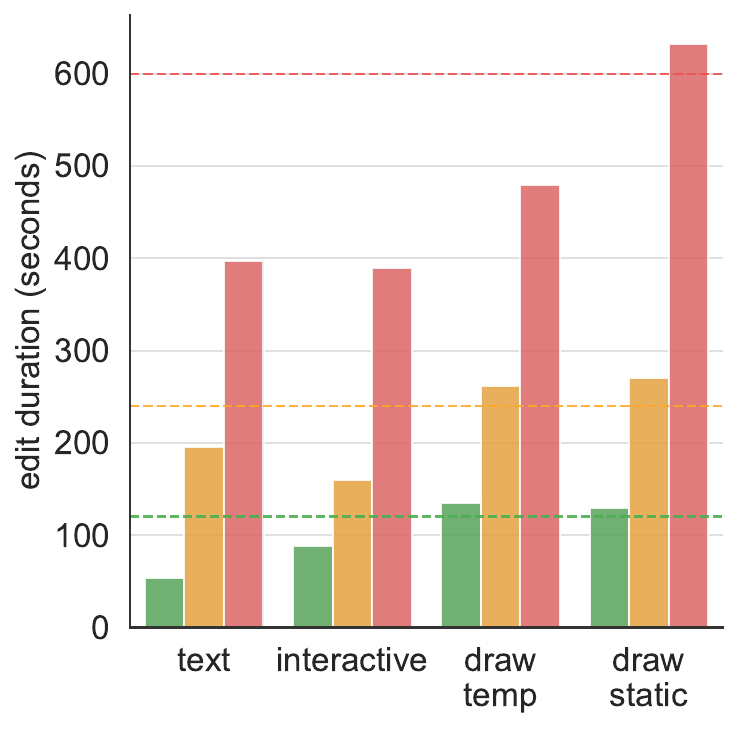}
        \caption{Duration of edit (dashed: guide times).}
        \label{fig:stats_edits_duration}
    \end{subfigure}
\hfill
    \begin{subfigure}{0.24\linewidth}
        \centering
        \includegraphics[height=82pt]{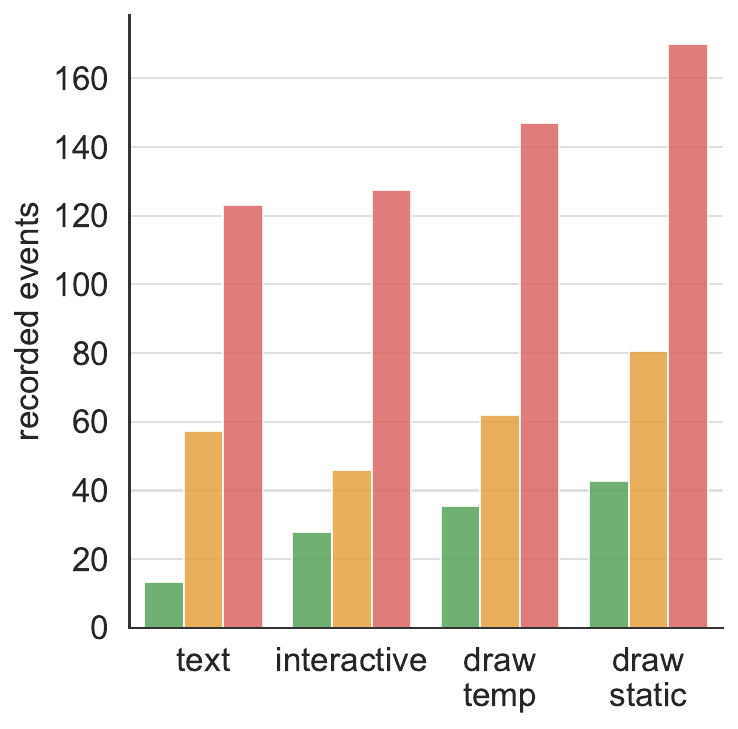}
        \caption{Number of events captured during edit.}
        \label{fig:stats_edits_events}
    \end{subfigure}
\vspace{-3pt}
\caption{
Request (a)-(b) and edit (c)-(d) statistics, grouped by modality (x-axis groups) and request difficulty: \easy{\bf easy}, \medium{\bf medium}, and \hard{\bf hard}.
}
\label{fig:stats}
\vspace{-10pt}
\end{figure}

\noindent\textbf{Requests.} \Cref{fig:stats_requests_words} shows that requests of each modality contained a similar number of words for easy requests. However, the non-text modalities include more words as the difficulties increase, suggesting that designers feel more comfortable communicating more complex instructions using speech. 
Despite having a lower number of words, \Cref{fig:stats_requests_duration} shows that typing takes longer than the other modalities. Note that the other modalities are additionally providing mouse gesturing and drawing in addition to speech. This provides strong evidence that multimodal naturalistic instructions provide a higher information density, and are better than typing-only for conditioning 3D edits.

\noindent\textbf{Edits.} \Cref{fig:stats_edits_duration} shows how long it took to perform edits instructed by easy, medium and hard requests. Recall requestors were given guide times of 2, 5 and 10 minutes (shows as dotted lines). 
While text and interactive edits took less time than requestors expected, requests that included drawing resulted in edits that were closer to the guide times. \Cref{fig:stats_edits_events} shows that edit procedures involving drawing contained more events. 
This suggests that the additional capability provided by the drawing tool allows for more complex and precise requests. As expected, the harder the request, the longer and more complex the edit.

\section{Experiments}
\vspace{-2pt}

\subsection{Baselines and AI models}
\vspace{-2pt}
There are no existing methods designed to perform multimodal-instructed 3D CAD editing, so we benchmark leading foundation models against human CAD experts:
\begin{itemize}
    \item \requestor{Ground truth (GT) human requestor.} The expert CAD designer who made the request. This can be considered a soft upper bound.
    \item \baseline{Human baseline.} An expert CAD designer, who did not create the request. 
    \item \gemini{Google Gemini 3 Pro.} Can natively support the request videos/audio.
    \item \gpt{OpenAI GPT 5.2.} No native video support, so images are extracted from the request video at 1FPS (the FPS Gemini samples at for fair comparison \cite{geminidocumentation}).
    \item \claude{Anthropic Claude Sonnet 4.5.} Chosen as it is a strong, fast coding model. Also has no native video support, so we follow the same protocol as for GPT.

\end{itemize}
All baselines and AIs are given the request video/audio, timestamped transcription and events recorded during the request, and the input CAD model. They are then asked to perform their edit. 
For Gemini, GPT and Claude, we implement a harness that allows the AI to import the input CAD .step file, write python CadQuery \cite{cadquery} scripts to modify the CAD model, run scripts with access to stdout and stderr, visually inspect the CAD model, update the scripts and re-run until it determines the task is completed.
Full access to CadQuery allows the AI to perform routines including geometry creation (including primitive creation, sketches, extrusions, revolutions), edge blending (including fillet and chamfer), boolean operations and assembly modelling.

\subsection{Metrics}

\noindent\textbf{Automatic.} These are standard 3D and feature metrics between the edit being evaluated and the human requestor ground truth edit. We use 3D \emph{Chamfer distance}, \emph{Voxel IoU}, and \emph{DINOv2 similarity} \cite{oquab2023dinov2} on rendered isometric views of CAD models.
Note that these metrics are less likely to handle ambiguity in the request, as the requestor's ground truth edit may not be the only possible solution. \emph{Validity} records the fraction of edits which produce a valid CAD model.

\noindent\textbf{Human eval.}
Instead of comparing to the human requestor ground truth edit, we instead show the request to a human CAD expert, alongside 6 orthographic views (top, bottom, front, back, left, right) and an isometric of the edit being evaluated. We ask the expert to give two scores between 1 (worst) and 7 (best). These are normalised to a 0-1 range. The first is \emph{instruction-following}, which rates how well the edit conforms to request. The second is \emph{model-quality}, which rates the quality of the work done. See the appendix for the rubrics used. Each edit is rated by 5 CAD experts\footnote{To ensure longevity of the human evaluation process, we are happy to run outputs produced by other researchers in the future through our pipeline.}, and we report the median for each score. The rubrics for both scores contain thresholds that indicate the output is of an acceptable quality to CAD experts ($5+$). We define \emph{acceptance} as the fraction of outputs that are above these thresholds for both scores. We consider acceptance as our primary metric, as if an AI cannot produce outputs that meet the high standards of designers, it will not be adopted.

\noindent\textbf{VLM eval.}
We follow the same protocol for human evaluation and use the same rubrics, but ask a VLM to grade using the same scales for instruction-following and model-quality. We present results from GPT 5.2 for our main VLM evaluations as we find it performs best as an evaluator. We acknowledge that there is a case where it evaluates itself, so we show the comparison of different VLM evaluators vs humans later in \Cref{sec:vlm_evaluators}.

Note that \benchmarkname\ (including the metrics used) is as agnostic to the editing method as possible. 
As long as an approach can load a B-Rep from a .step file, take a multimodal request, and export a .step file, it can be evaluated. By keeping the benchmark general, we will not be limiting research to a specific class of editing methods, instead leaving the door open for researchers to experiment with a wide range of approaches.

\subsection{Results}

\begin{table}[t]
\centering
\caption{Main results. For chamfer distance, lower is better. Higher is better for everything else. All AI approaches perform worse on all metrics than the human baseline.}
\vspace{-5pt}
\resizebox{\textwidth}{!}{
\begin{tabular}{@{}lllllllllll@{}}
\toprule
                                          & \multicolumn{4}{c}{Automatic metrics}                                                                     & \multicolumn{3}{c}{VLM eval}                                                            & \multicolumn{3}{c}{Human eval}                                                          \\ 
\cmidrule(lr){2-5}
\cmidrule(lr){6-8}
\cmidrule(lr){9-11}
                                          & Chamfer-dist             & Voxel-IoU                & DINO-sim                 & Validity                 & Instruction                 & Quality                     & Acceptance                  & Instruction                 & Quality                     & Acceptance                  \\
\requestor{GT Human requestor} & {\color[HTML]{D0D0D0} -} & {\color[HTML]{D0D0D0} -} & {\color[HTML]{D0D0D0} -} & {\color[HTML]{D0D0D0} -} & {\color[HTML]{D0D0D0} 0.56} & {\color[HTML]{D0D0D0} 0.70} & {\color[HTML]{D0D0D0} 0.47} & {\color[HTML]{D0D0D0} 0.74} & {\color[HTML]{D0D0D0} 0.66} & {\color[HTML]{D0D0D0} 0.82} \\
\baseline{Human baseline}                            & \textbf{22}                       & \textbf{0.76}            & \textbf{0.93}            & \textbf{1.00}            & \textbf{0.55}               & \textbf{0.70}               & \textbf{0.45}               & \textbf{0.74}               & \textbf{0.66}               & \textbf{0.78}               \\
\claude{Claude Sonnet 4.5}                         & 54                       & 0.18                     & 0.25                     & 0.42                     & 0.35                        & 0.50                        & 0.15                        & 0.22                        & 0.10                        & 0.05                        \\
\gemini{Gemini-3-Pro}                              & 110                      & 0.30                     & 0.36                     & 0.58                     & 0.40                        & 0.56                        & 0.19                        & 0.27                        & 0.16                        & 0.10                        \\
\gpt{GPT 5.2}                                   & 50                       & 0.57                     & 0.66                     & 0.99                     & 0.50                        & 0.66                        & 0.30                        & 0.48                        & 0.39                        & 0.25                        \\ \bottomrule
\end{tabular}
}
\label{tab:main_results}
\vspace{-10pt}
\end{table}

Table \ref{tab:main_results} shows results on the whole benchmark with automatic, VLM and human evals.
Note that we do not report automatic metrics for the GT human requestor, as automatic metrics compare against the requestor's edit. We provide VLM and human ratings for these ground truth edits as a reference, but they are grayed out, as the fairest human comparison is the human baseline.
Claude, Gemini and GPT all performed much worse than the human baseline across all metrics, most notably by at least 50\% on human rater acceptance. Every metric gave the same ordering, from best to worst, as \baseline{Human baseline}, \gpt{GPT}, \gemini{Gemini}, \claude{Claude}, apart from chamfer distance, for which Claude and Gemini swap places.

Whilst VLM rating did give the same ordering as human ratings, it was less able to identify high quality human edits than human raters (0.45 vs 0.78 acceptance). It also over-estimated the quality of AI edits compared to how humans rated (\eg 0.50 vs 0.10 when rating Claude).
An interesting finding is that human instruction, quality and acceptance scores are around 0.7. This is due to (1) the rubric being strict, with higher scores of 6$+$/7 being reserved for outstanding work produced by an especially helpful editor going ``above and beyond'', and 
(2) the human evaluators may be interpreting any ambiguity in the requests differently. Nevertheless, the GT human requestor and human baseline being scored so closely (same instruction and quality scores, and 0.78 and 0.82 acceptance scores) indicates the human evaluation is robust, and has headroom to capture AI outperforming humans in the future.

Surprisingly, although Gemini ranks higher than GPT and Claude on visual understanding benchmarks \cite{wang2024mmlu}, GPT performs better on \benchmarkname. Factors that may have contributed include Gemini
being less persistent on average (6.8 vs 8.1 script iterations), and generating fewer tokens on average arriving at the final CAD model (191K vs 504K combined thinking and output).
This highlights a strength of \benchmarkname. For an AI to do well, it must posses three capabilities that are usually benchmarked separately, yet are all considered fundamental to general machine intelligence: multimodal-understanding \cite{patraucean2023perception}, coding \cite{jimenez2023swe} and 3D reasoning/manipulation \cite{mu2021maniskill}.

\begin{figure}
    \centering
    \includegraphics[width=1\linewidth]{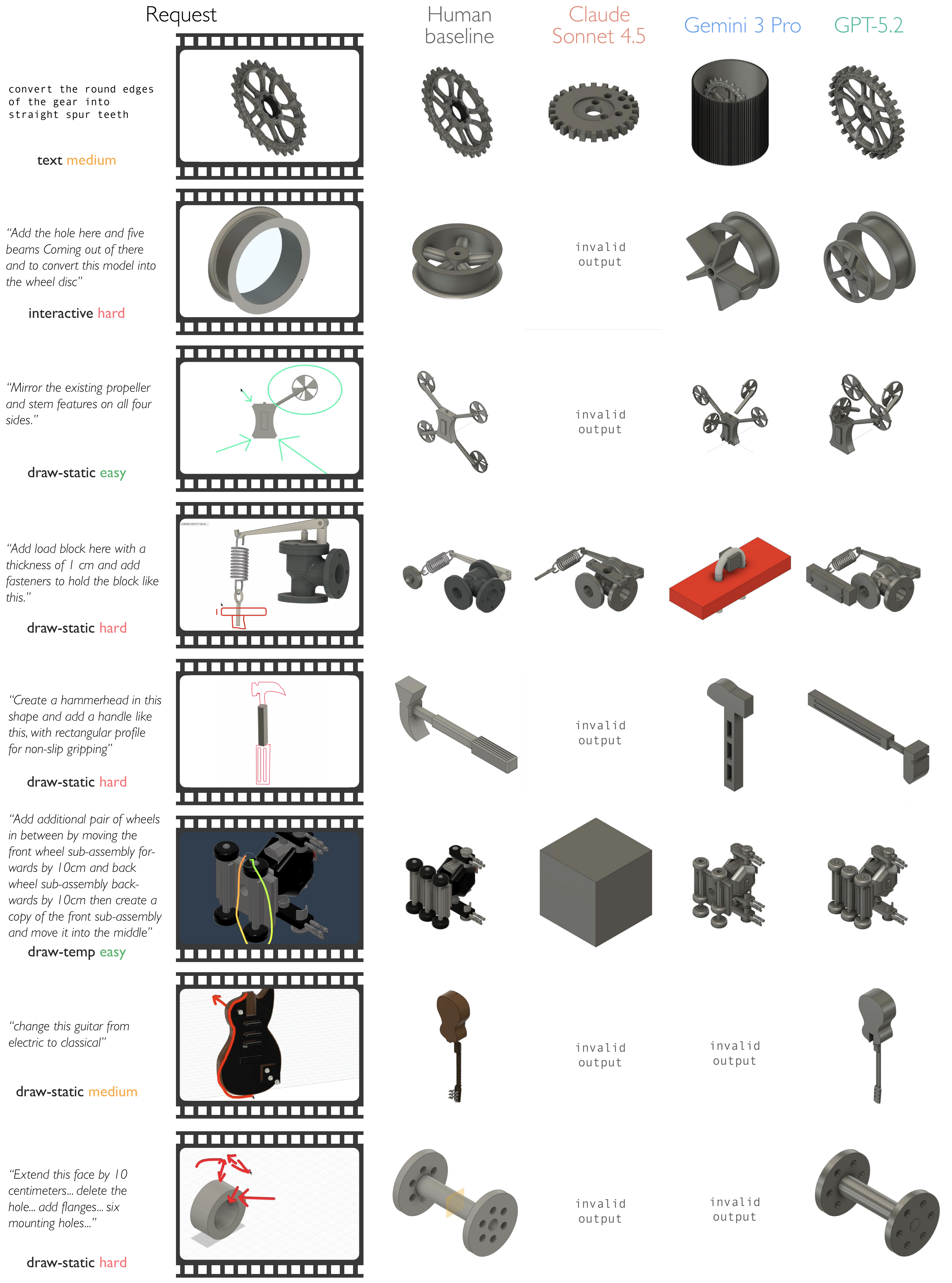}
    \caption{Qualitative examples of edits performed by humans and AIs. Models particularly struggled with compositional visual reasoning, resulting in mistakes like adding spokes that don't connect to the main assembly, or adding propellers at the wrong set of corners. GPT 5.2 shows the most promise, successfully performing the edits in the bottom four rows. Video examples provided on the project webpage.}
    \label{fig:qualitative-examples}
\end{figure}

\Cref{fig:qualitative-examples} shows example outputs.
For the drone (row 3), Gemini and GPT are both able to probe the model, identify the rotors and arms and copy them the correct number of times. However, they are unable to see that they have been misplaced, declaring the edit as finished. This suggests that nuanced visual reasoning, including long-standing problems such as relative 3D positioning \cite{johnson2017clevr} and compositionality \cite{thrush2022winoground}, remain unsolved. For the gear, Claude gave up trying to perform the edit, and instead wrote a script to generate from scratch, suggesting limited ability to understand the existing 3D model. Falling back to solid generation rather than editing is not a surprise, as it is a task for which there is a larger amount of public data to train on and memorize \cite{Koch_2019_CVPR}. Impressively, GPT was able to understand the request in the bottom row, associating words with the correct arrows as they are drawn, and successfully edited the model (video provided on the project webpage).

\subsection{Results analysis}

\begin{figure}[t]
    \centering
    \begin{subfigure}{0.49\linewidth}
        \centering
        \includegraphics[height=80pt, trim = 0 0 0 0]{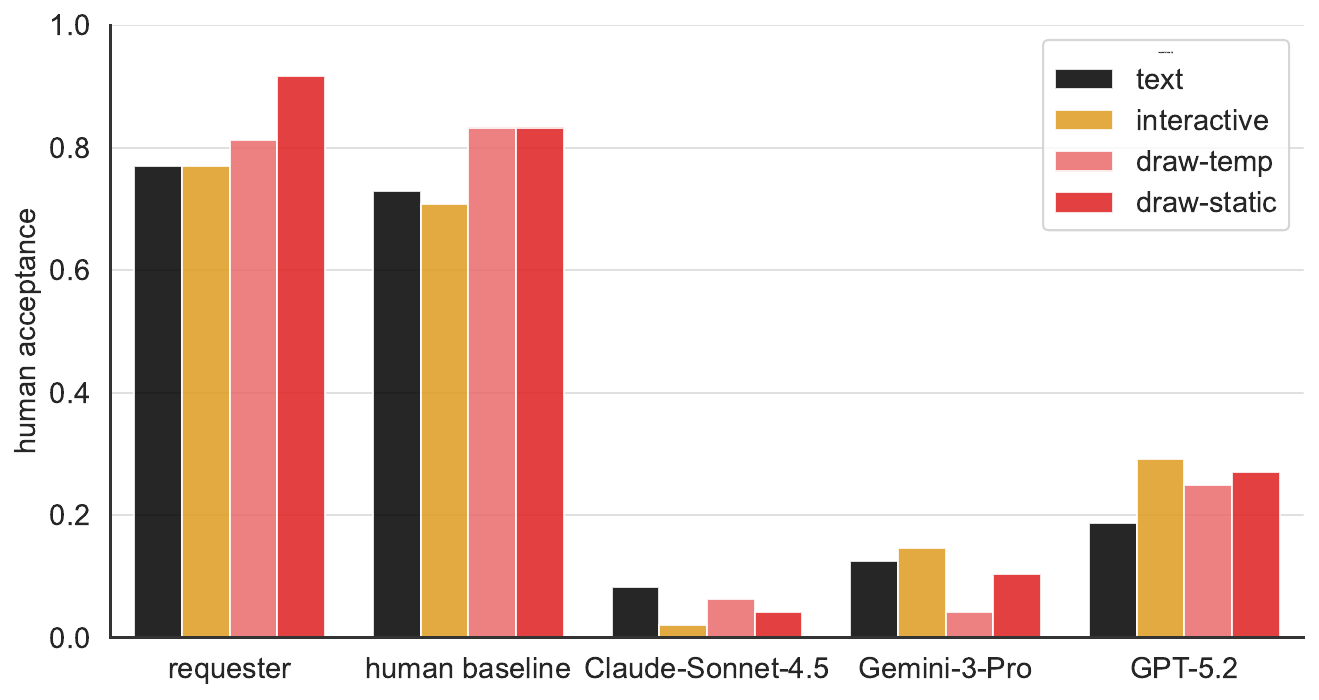}
        \caption{Modality}
        \label{fig:results-modality}
    \end{subfigure}
\hfill
    \begin{subfigure}{0.49\linewidth}
        \centering
        \includegraphics[height=80pt, trim = 0 0 0 0]{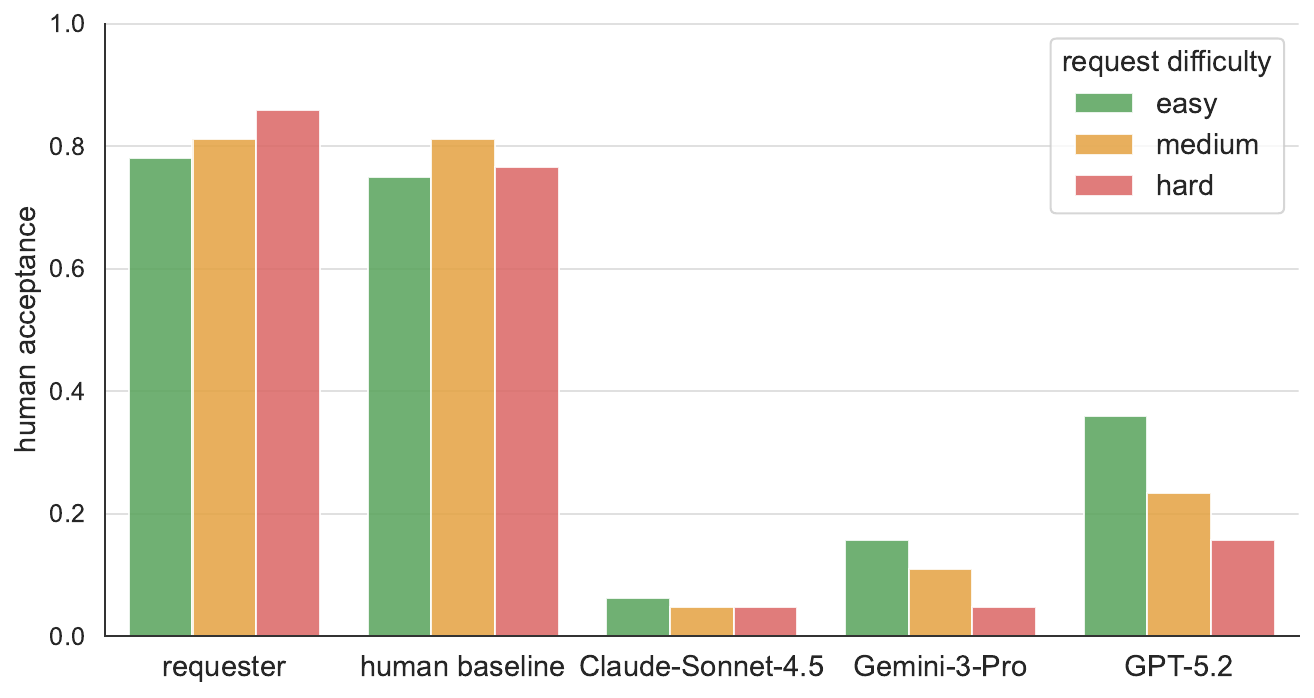}
        \caption{Difficulty}
        \label{fig:results-difficulty}
    \end{subfigure}
\vspace{-3pt}
\caption{Acceptance scores for all models, broken down by modality and difficulty.}
\vspace{-5pt}
\label{fig:results-breakdown}
\end{figure}
\noindent\textbf{Modality.}
\Cref{fig:results-modality} shows the acceptance score of all baselines and AIs, broken down by modality. Edits performed by humans on requests which included drawing are scored highest, as humans (both editors and evaluators) are able to interpret the extra information and precision indicated by drawings.
However, this was not the case for Claude, Gemini or GPT, which suggests there is room for improvement in their multimodal understanding, in particular simultaneous talking and drawing.

\noindent\textbf{Difficulty.}
\Cref{fig:results-difficulty} shows the acceptance scores, broken down by difficulty. The human baseline performs similarly well on all difficulties. GPT and Claude perform better when given easier requests. This suggests that measuring performance on the different difficulty levels in \benchmarkname\ will be a good way of tracking AI improvement over time.

\begin{figure}[t]
    \centering
    \begin{subfigure}{0.49\linewidth}
        \includegraphics[height=210pt, trim = 0 0 0 0]{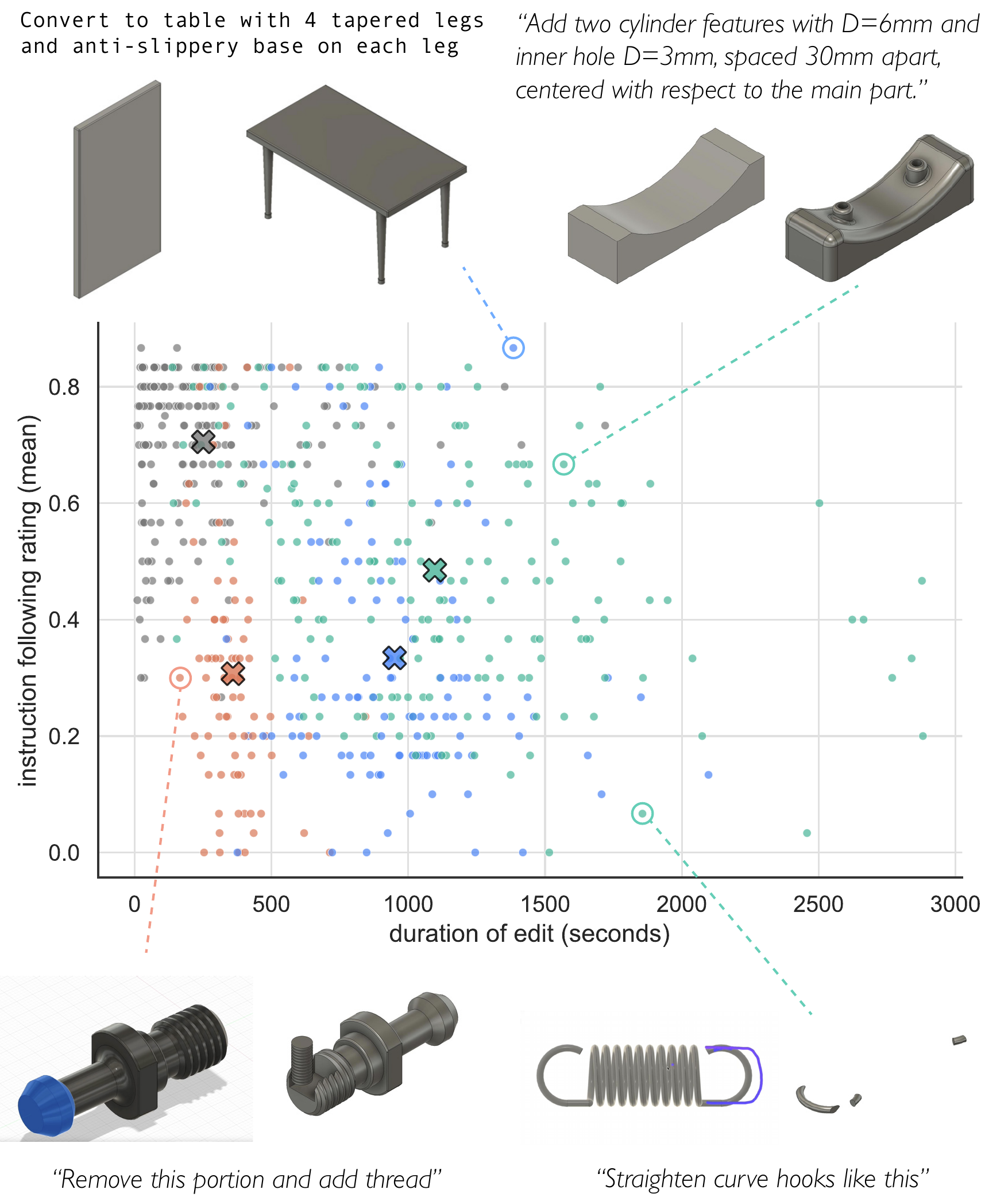}
        \caption{Time vs score.}
        \label{fig:speed_accuracy}
    \end{subfigure}
\hfill
    \begin{subfigure}{0.49\linewidth}
        \includegraphics[height=210pt, trim = 0 0 0 0]{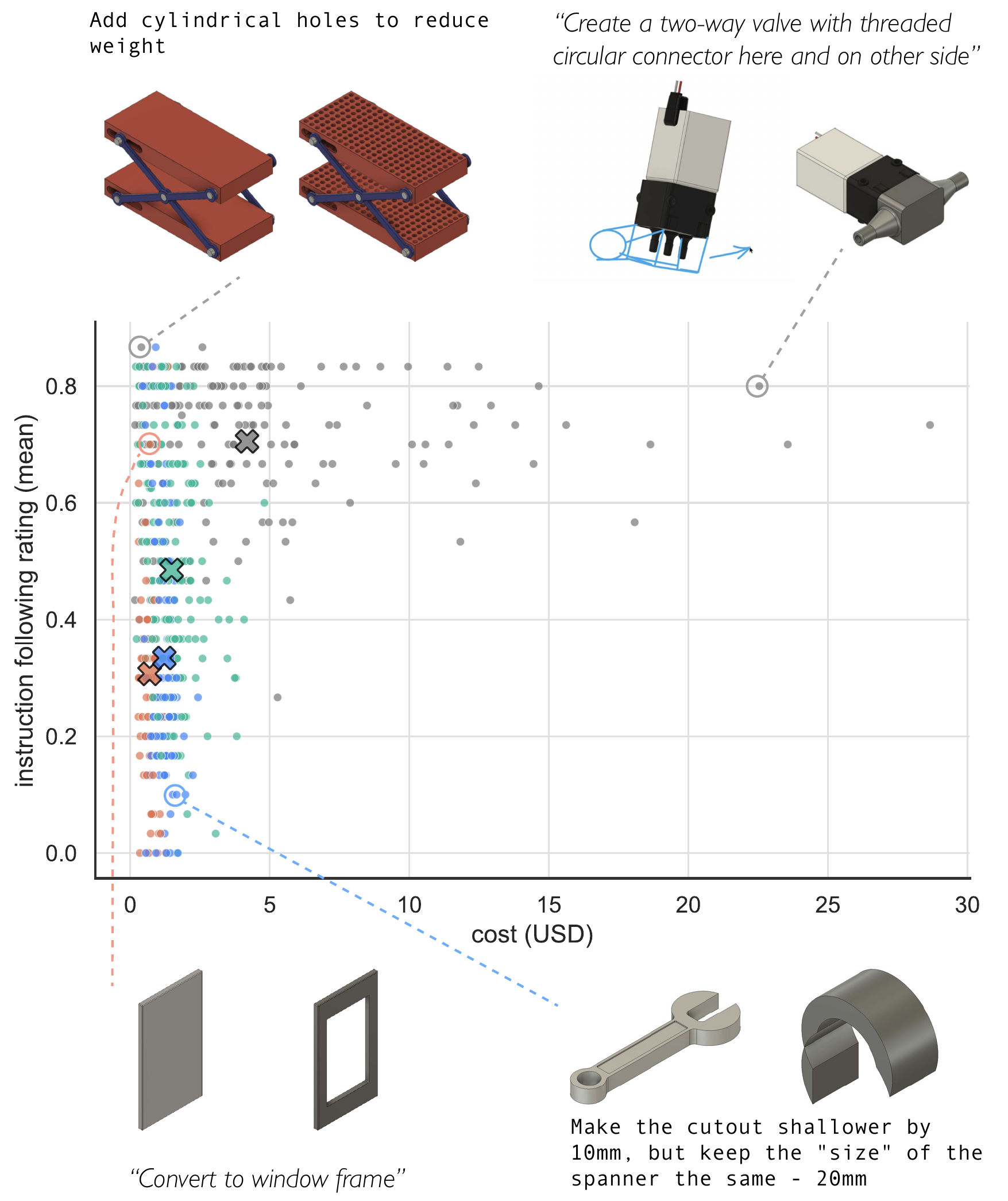}
        \caption{Cost vs score. Humans rated at \$60 an hour.}
        \label{fig:performance_by_cost}
    \end{subfigure}
    \caption{Efficiency of \baseline{humans}, \claude{Claude}, \gemini{Gemini} and \gpt{GPT}. Averages are shown as crosses. Humans complete their edits faster, but cost more. Illustrative pairs of requests (left images) and edits (right images) are highlighted to show range of edit qualities. Text is high-level summary of the request.}
    \vspace{-5pt}
\label{fig:results_time}
\end{figure}

\begin{figure}[t]
    \centering
    \includegraphics[width=1\linewidth]{figures/distribution_of_edit_duration_by_difficulty_by_model.pdf}
    \vspace{-20pt}
    \caption{Time taken to perform \easy{easy}, \medium{medium} and \hard{hard} edits. Humans clearly spent longer editing the harder the request, whereas this effect is much less pronounced for the best AI (GPT), and not observed for Gemini and Claude.}
    \label{fig:duration_by_difficulty_and_model}
    \vspace{-10pt}
\end{figure}

\noindent\textbf{Efficiency.}
Here, we analyse the time and cost of humans and AIs performing edits. \Cref{fig:speed_accuracy} shows the time vs score of the {human baseline} and {Claude}, {Gemini} and {GPT}. Humans are faster and produce better work. \Cref{fig:performance_by_cost} shows that, when adjusting for cost (rating humans at \$60 per hour), the higher quality outputs from humans are reflected in a higher cost per edit.

\Cref{fig:duration_by_difficulty_and_model} shows how long it takes humans and AIs to complete {easy}, {medium} and {hard} edits. As expected, humans clearly spend longer on an edit the more difficult the request. Whilst this behavior exists for GPT (the best AI), it is much less pronounced, and non-existent for Gemini and Claude. Note that there is ample opportunity for runtime to vary, both through internal reasoning mechanisms and the harness we provide. Although GPT did use more tokens to arrive at its solutions overall, this suggests that environment- and task-aware test-time scaling is an open problem for large-scale foundation models.

\subsection{Metrics analysis}

\begin{figure}[t!]
    \begin{minipage}[t]{0.48\textwidth}
        \centering
        \includegraphics[height=95pt]{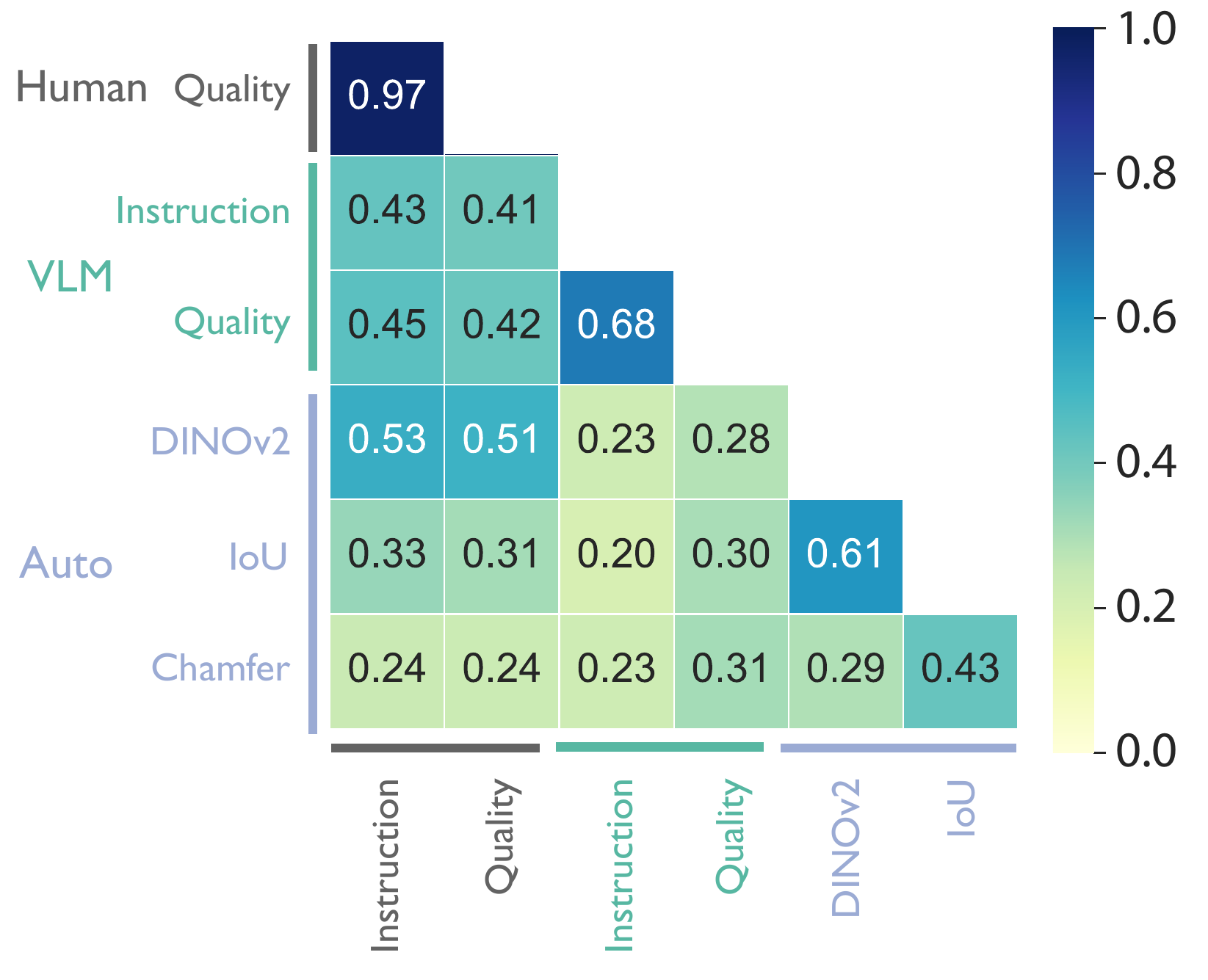}
        \vspace{-5pt}
        \caption{Spearman rank correlations between human ratings, VLM ratings (GPT 5.2), and automatic metrics.}
        \label{fig:metric_correlation}
    \end{minipage}\hfill
    \begin{minipage}[t]{0.48\textwidth}
        \centering
        \includegraphics[height=100pt]{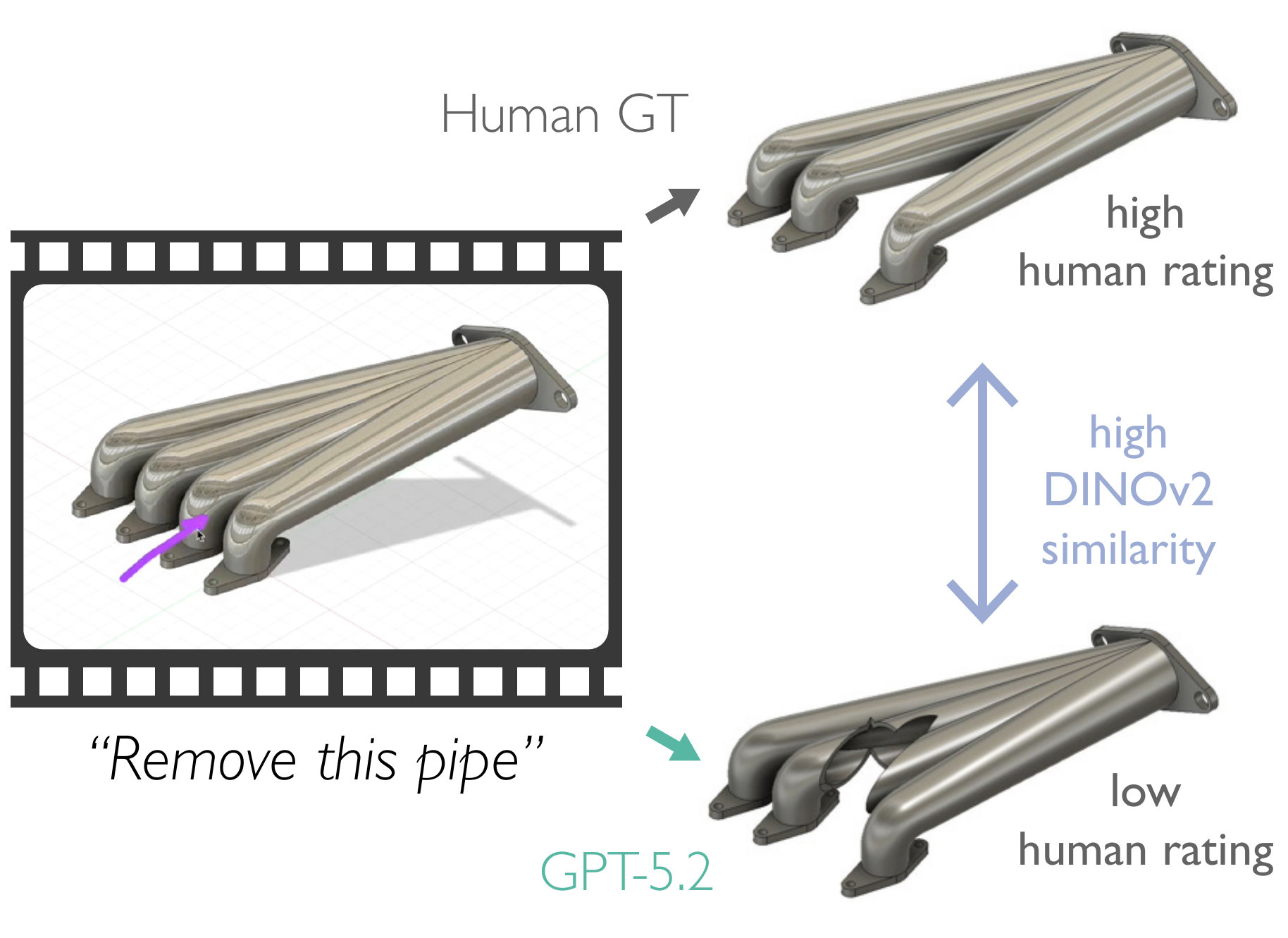}
        \vspace{-5pt}
        \caption{Two edits rated high and low by CAD experts. In this case, visual similarity (DINOv2) differentiates poorly.}
        \label{fig:metric_examples}
    \end{minipage}
    \vspace{-10pt}
\end{figure}

\noindent\textbf{Human rater reliability.}
To quantify inter-rater reliability, we compute the two-way random-effects Intraclass Correlation Coefficient (ICC) with absolute agreement \cite{koo2016guideline}. 
Importantly, the reliability of the mean of five raters is high (ICC$(2,k) = 0.88$, 95\% CI: $[0.87, 0.89]$), indicating that the aggregated human scores provide a stable and consistent estimate of edit quality.

\noindent\textbf{Human vs VLM vs automatic metrics.}
To determine how closely VLM ratings and automatic metrics follow human judgments, we compute Spearman Rank Correlations between all evaluations and metrics (\Cref{fig:metric_correlation}).
Human ratings of instruction following and quality are strongly correlated, indicating substantial overlap between successful instruction execution and perceived edit quality.
VLM ratings of instruction following and quality are less well correlated and show limited correlation with human ratings.
This limited alignment indicates that current VLM-based evaluators do not reliably capture the factors that humans consider when assessing the quality of CAD edits.
This is consistent with our observation that VLMs struggle to produce high quality edits themselves, and may reflect limitations in their visual understanding and ability to evaluate progress towards their goals.
Of all the automatic metrics explored, although DINOv2 similarity sometimes fails to identify low-quality edits that are visually similar to the ground truth (\Cref{fig:metric_examples}), it is the metric most correlated with human ratings, suggesting that visual feature similarity provides the most informative automatic proxy signal to human ratings.

\begin{table}[b]
\vspace{-5pt}
\centering
\caption{Humans and AIs rating each others' edits.}
\vspace{-3pt}
\resizebox{\textwidth}{!}{
\begin{tabular}{@{}lllllllllllll@{}}
\toprule
                  & \multicolumn{3}{c}{\baseline{Human evaluator}} & \multicolumn{3}{c}{\claude{Claude Sonnet 4.5 evaluator}} & \multicolumn{3}{c}{\gemini{Gemini 3 Pro evaluator}}   & \multicolumn{3}{c}{\gpt{GPT 5.2 evaluator}}                  \\ 
\cmidrule(lr){2-4}
\cmidrule(lr){5-7}
\cmidrule(lr){8-10}
\cmidrule(lr){11-13}
                  & Instruction  & Quality  & Acceptance  & Instruction & Quality & Acceptance & Instruction & Quality & Acceptance & Instruction & Quality & Acceptance \\
\baseline{Human baseline}    & 0.74        & 0.66    & 0.78       & 0.57         & 0.65     & 0.61        & 0.85        & 0.95    & 0.99       & 0.55        & 0.70    & 0.45       \\
\claude{Claude Sonnet 4.5} & 0.22        & 0.10    & 0.05       & 0.49         & 0.55     & 0.53        & 0.55        & 0.64    & 0.64       & 0.35        & 0.50    & 0.15       \\
\gemini{Gemini 3 Pro}      & 0.27        & 0.16    & 0.10       & 0.50         & 0.56     & 0.54        & 0.63        & 0.74    & 0.73       & 0.40        & 0.56    & 0.19       \\
\gpt{GPT 5.2}           & 0.48        & 0.39    & 0.25       & 0.48         & 0.55     & 0.49        & 0.79        & 0.89    & 0.93       & 0.50        & 0.66    & 0.30        \\ \bottomrule
\end{tabular}
}
\label{tab:vlm_evals}
\end{table}

\noindent\textbf{Comparison of VLMs as evaluators.}\label{sec:vlm_evaluators} For our main experiments in \Cref{tab:main_results}, we presented results using GPT 5.2 as the VLM evaluator as we find it performs best. \Cref{tab:vlm_evals} presents VLM evaluations from all the AIs (Claude, Gemini and GPT) and humans, when all rating each other. All agree humans are the best editors. Gemini and GPT both agree with the human rankings, although Gemini has a very high acceptance. Interestingly, Claude ranks itself above GPT. However, we already know it performs edits poorly, and its lack of visual understanding likely causes it to be a poor evaluator as well. All models give a much narrower margin between human and AI edits, suggesting current VLMs may be less reliable as differentiators as AI edits improve towards human levels.

\vspace{-2pt}
\subsection{Conclusion}
\vspace{-2pt}

In this paper, we introduced \benchmarkname, the first benchmark for 3D CAD editing collected from expert CAD users. \benchmarkname\ is the first to capture multimodal editing requests, including video, speech and drawings, all with full interactivity with CAD models in CAD software.
We analyzed the requests and edits performed by CAD experts, and found that multi-modal requests convey more information in a shorter time than text-only, and elicit more complex edits.
We benchmarked the ability of frontier models to understand multimodal requests and perform 3D edits against the standard set by human CAD experts, finding that even the most capable model (GPT 5.2) is far below human ability.

As well as providing a solid benchmark that 3D CAD editing and foundation models can be developed around, \benchmarkname\ motivates a number of future research directions. These include training models for editing, designing systems for and evaluating humans and AI working collaboratively on design tasks, and exploring the editing process in addition to the final output.

\subsection*{Acknowledgements}
We would like to thank Ciklum (in particular Martin Valko, Marek Mikles, Svitlana Bleichyk and Iuliana Prykhodko) for their CAD expertise and coordination. We would like to thank the following from Autodesk for their feedback on the project: 
Sean J. Liu,
Joseph Lambourne,
Long Zhang
Justin Matejka,
Yilin Liu,
Joanne Leong,
John Thompson,
Victor Oliveira Antonino,
Tonya Trapp,
Mike Haley.

\bibliographystyle{splncs04}
\bibliography{main}

\appendix

\section{Video examples}
The videos on the project webpage contain four examples from \benchmarkname. One request from each of the four modalities is shown. They are described in Section 2.2 of the main paper. To recap, they are:

\begin{itemize}
\item \textbf{Text}. This is the simplest form of request, collected by typing into a text box with no model manipulation or speech. 
\item \textbf{Interactive}. Video of the requestor interacting with the CAD model, gesturing/pointing with the mouse, and saying their instructions out loud concurrently.
\item \textbf{Draw-temp}. As interactive, but the requestor also uses a drawing tool to help illustrate the request. Drawings disappear after a few seconds.
\item \textbf{Draw-static}. As for draw-temp, but drawings remain visible for the whole request or until manually deleted.
\end{itemize}

\noindent Note that for text, although the video is displayed to show typing speed, we pass the raw text string into Claude, Gemini and GPT so they do not have to perform OCR. 

\noindent In the video, the example modalities and difficulties, in order, are:
\begin{enumerate}
    \item Draw-static, hard
    \item Draw-temp, easy
    \item Interactive, hard
    \item Text, medium
\end{enumerate}

\noindent For each modality, we show:
\begin{enumerate}
    \item Request video.
    \item Sped up viewport renders from the requestor performing the edit, and the human baseline (who has only seen the request video) performing the edit. Logged events are displayed as captions.
    \item Output CAD models from humans and AIs alongside the input model.
\end{enumerate}

\section{Rubrics}
The human and VLM evals use the same rubrics for rating edits. We collect two scores from each: \emph{instruction following}, which measures how well the edit conforms to the request, and \emph{quality}, which measures the quality of the work done. Both are on a 1-7 scale, from worst to best. We list both rubrics here:

\subsubsection{Instruction following}
\begin{enumerate}
    \item Makes things worse or goes in completely the wrong direction.
    \item Does nothing.
    \item Makes an attempt, demonstrates a rough understanding of the request but many parts incorrect or incomplete.
    \item Mostly does what is asked, but with noticeable errors or omissions.
    \item Does what is asked, with small errors or omissions.
    \item Perfect - follows the request precisely with no errors or omissions.
    \item Above and beyond - perfect with helpful extras that a thoughtful expert designer would include.
\end{enumerate}

\subsubsection{Quality}
\begin{enumerate}
    \item No model or edit produced.
    \item Very poor - erroneous model, impossible geometry, overlapping geometries, incomplete sketches etc.
    \item Poor - simplistic, blocky.
    \item Average - acceptable first pass.
    \item Good attempt with room for improvement.
    \item High quality work, as you’d expect from an experienced designer.
    \item Perfect - extremely polished design.
\end{enumerate}

\section{Filtering criteria for input CAD model selection}

We filter models from the Fusion Gallery dataset into 4 distinct categories, with the following criteria:
\begin{itemize}
\item \textbf{Parametric assemblies} contain between 5 and 200 parametric operations in the construction timeline and at between 2 and 30 components/bodies.
\item \textbf{Parametric single-bodies} contain between 5 and 200 parametric operations in the construction timeline and exactly 1 component/body.
\item \textbf{Dumb assemblies} contain 0 parametric operations in the construction timeline and between 2 and 30 components/bodies.
\item \textbf{Dumb single-bodies} contain 0 parametric operations in the construction timeline and exactly 1 component/body. As there were few such bodies in the Fusion Gallery Dataset itself, we also extracted individual components from the Fusion Gallery Assemblies dataset and removed the construction timeline.
\end{itemize}
We then visually inspected candidates to select a diverse range of input models. Input models have an average of 298 faces, 782 edges and 504 vertices.

\end{document}